\documentclass[runningheads]{llncs}

\usepackage{amssymb}
\usepackage{graphicx}
\usepackage{subfigure}
\usepackage{dsfont}
\usepackage[ruled, norelsize]{algorithm2e}
\usepackage{times}

\usepackage{url}
\urldef{\mailsa}\path|mewcheng@gmail.com, liuzunren@126.com|
\urldef{\mailsb}\path|hongwei_z@cmbchina.com|
\urldef{\mailsc}\path|act@uow.edu.au|
\newcommand{\keywords}[1]{\par\addvspace\baselineskip
\noindent\keywordname\enspace\ignorespaces#1}

\begin{document}

\mainmatter  % start of an individual contribution

% first the title is needed
\title{A Family of Maximum Margin Criterion for Adaptive Learning}

% a short form should be given in case it is too long for the running head
%\titlerunning{Lecture Notes in Computer Science: Authors' Instructions}

% the name(s) of the author(s) follow(s) next
%
% NB: Chinese authors should write their first names(s) in front of
% their surnames. This ensures that the names appear correctly in
% the running heads and the author index.
%
%\author{Miao Cheng$ ^\dag $
%%\thanks{Please note that the LNCS Editorial assumes that all authors have used
%%the western naming convention, with given names preceding surnames. This determines
%%the structure of the names in the running heads and the author index.}%
%\and Zunren Liu$ ^\dag $ \and Hongwei Zou$ ^\ddag $ \and Ah Chung Tsoi$ ^\S $ }
%%\and\\
%%Anna Kramer\and Leonie Kunz\and Christine Rei\ss\and\\
%%Nicole Sator\and Erika Siebert-Cole\and Peter Stra\ss er}
%%
%\authorrunning{Miao Cheng, Zunren Liu et al.}
%% (feature abused for this document to repeat the title also on left hand pages)
%
%% the affiliations are given next; don't give your e-mail address
%% unless you accept that it will be published
%\institute{ $ ^\dag $Department of Computer Science, Qingdao University, Qingdao, China	\\
%$ ^\ddag $Division of Information Technology, Chongqing Branch of China Merchants Bank, Chongqing, China	\\
%$ ^\S $Department of Computer Science and Software Engineering, University of Wollongong, Australia	\\
%\mailsa\\
%\mailsb\\
%%\mailsc\\
%%\url{http://www.springer.com/lncs}}

\author{Miao Cheng\inst{1}
%\thanks{Please note that the LNCS Editorial assumes that all authors have used
%the western naming convention, with given names preceding surnames. This determines
%the structure of the names in the running heads and the author index.}%
\and Zunren Liu\inst{1} \and Hongwei Zou\inst{2} \and Ah Chung Tsoi\inst{3} }
%\and\\
%Anna Kramer\and Leonie Kunz\and Christine Rei\ss\and\\
%Nicole Sator\and Erika Siebert-Cole\and Peter Stra\ss er}
%
\authorrunning{Miao Cheng, Zunren Liu et al.}
% (feature abused for this document to repeat the title also on left hand pages)
% the affiliations are given next; don't give your e-mail address
% unless you accept that it will be published
\institute{ Department of Computer Science, Qingdao University, Qingdao, China	\\
\mailsa	\and
Division of Information Technology, Chongqing Branch of China Merchants Bank, Chongqing, China	\\
\mailsb	\and
Department of Computer Science and Software Engineering, University of Wollongong, Australia	\\
\mailsc
}

%
% NB: a more complex sample for affiliations and the mapping to the
% corresponding authors can be found in the file "llncs.dem"
% (search for the string "\mainmatter" where a contribution starts).
% "llncs.dem" accompanies the document class "llncs.cls".
%

%\toctitle{Lecture Notes in Computer Science}
%\tocauthor{Authors' Instructions}
\maketitle

\begin{abstract}
In recent years, pattern analysis plays an important role in data mining and recognition, and many variants have been proposed to handle complicated scenarios. In the literature, it has been quite familiar with high dimensionality of data samples, but either such characteristics or large data sets have become usual sense in real-world applications. In this work, 
%the statistical linear methods are re-visited, and 
an improved maximum margin criterion (MMC) method is introduced firstly. With the new definition of MMC, several variants of MMC, including random MMC, layered MMC, 2D$ ^2 $ MMC, are designed to make adaptive learning applicable.
%More specifically, a random MMC is designed to alleviate the intrinsic limitations of existing methods. Furthermore, the layered MMC and 2D MMC are developed to meet demands of different data categories. In light of ideas of neural network structures, the MMC Network is also proposed to learn deep features of images. 
Particularly, the MMC network is developed to learn deep features of images in light of simple deep networks.
Experimental results on a diversity of data sets demonstrate the discriminant ability of proposed MMC methods are compenent to be adopted in complicated application scenarios. 
%\emph{abstract} environment. 
\end{abstract}

% *************************************************************************************************************
\section{Introduction}
As a promising step, feature extraction has become an important approach to data mining and pattern recognition. 
%Until now, there have been many excellent solutions proposed to diverse applications with efficient performance. 
%Among these methods, subspace learning methods are competent to learn representative patterns with stationary results in theory. 
%Nevertheless,
And traditional methods usually suffer from intrinsic limitations from characteristics of original data. The first one refers to high-dimensionality of samples that hinders efficient calcualtion, and the outstanding solutions come down to direct approach to scatter matrices decomposition. %while informative subspace is preserved with discriminant power. 
Furthermore, there arise broad interests in large-scale data mining in many real-world applications, and
%which requires affordable method for fast large data processing. As a result, the sample quantity become large as well as calculational complexity, if %tranditional solutions are considered. 
this pushes new challege for feature analysis and reduction. In terms of such demands, it has become a vivid research topic to devise improved learning methods to conduct high-dimensional data with large amount meanwhile.
%\cite{Sakurai02SIRA}\cite{Tang16LSHD}\cite{Gudmundsson10}.

In the literature, principal component analysis (PCA)~\cite{Turk91PCA} and linear discriminant analysis (LDA)~\cite{Belhumeur97LDA}~\cite{Bishop11PRML} have become popular methods for pattern analysis in statistical learning theory. 
%The main differences between those two methods rely on the fact that PCA focuses on unsupervised learning that separates whole data with centering %transformation. On the contrary, LDA aims to maximize the between-class scatter $ S_b $ while minimize the within-calss scatter $ S_w $ simultaneously. %And the intrinsic limitation of original LDA is also related with ratio formalism of discriminant scatters, which is handled with stepwise calculations of %scatter decomposition \cite{Yu01DLDA}. 
To address high-dimensional problem of original data, there has been a common sense that reaches the eigen-decomposition of scatter matrices with calculational tricks of sub-matrices' multiplications~\cite{Yu01DLDA}.
%, e.g., $ S_bv = \lambda S_w v $. 
Besides traditional ratio LDA, there is another approach to do discriminant analysis with subtraction formalism, i.g., maximum margin criterion (MMC), while null denominator problem can be avoided~\cite{Li06MMC}~\cite{Liu07MMC}. Nevertheless, just like an old Chinese saying goes, "There would be something else in loss if something was obtained". The calculational trick of sub-matrices is unavailable for MMC anymore, and extra calculations are to be invovled in general. In the previous work of ours, a direct solution is proposed to handle the calculational limitation of large discriminant scatter of high-dimensional data for MMC~\cite{Cheng11DMMC}. And discriminant analysis can be proceeded straightforward while both $ S_b $ and $ S_w $ scatters are considered together with preserved efficiency.

In this work, the direct MMC framework is further developed to conduct adaptive learning, and insensitive to high-dimensionality problem of data in any scenarios. 
%In addition, a random mechnism is proposed to deal with such dilemma that data quantity is too large to adopt direct calculations. In other words, the %proposed method can be regarded as a generalized discriminant approach to ubiquitous subspace learning.
Furthermore, several extensions of MMC are proposed to conduct adaptive classification of different categories of data.
The rest of this paper is organized as follows. The background knowledge of direct MMC is reviewed in Section 2, while the calculation efficiency is discussed in theory, followed by the details of an improved MMC in Section 3. And then, several extensions of MMC is proposed for applications of different scenarios. A set of comparison experiments on discriminant learning are given in Section 4. Finally, the conclusion is draw in Section 5.

\section{Direct Maximum Margin Criterion}
The original MMC considers the substraction of discriminant scatters of original data. Given data set $ X = \left[ {{x_1},{x_2}, \cdots ,{x_n}} \right] \in {R^{d \times n}} $ in $ c $ classes, the $ S_b $ and $ S_w $ scatters are defined as
\begin{equation}
 \begin{array}{l}
{S_b} = \sum\limits_{i = 1}^c {{n_i}\left( {{m_i} - m} \right){{\left( {{m_i} - m} \right)}^T}} = X{L_b}{X^T} \\ 
{S_w} = \sum\limits_{i = 1}^c {\sum\limits_{j = 1}^{{n_i}} {\left( {{x_{ij}} - {m_i}} \right){{\left( {{x_{ij}} - {m_i}} \right)}^T}} } = X{L_w}{X^T} 
\end{array} 
\end{equation} 
where $ m_i $ and $ m $ respectively denote the mean data of $ i $-th class and whole data, and $ n_i $ denotes the sample amount of $ i $-th class. Besides, the disrciminant scatters can also be desceibed in graph formula with definition of Laplacian matrices $ L_b $ and $ L_w $ \cite{Cheng10IEL}\cite{Yan07GE}. As a result, MMC solves the following quadratic optimization objective to find the ideal $ w $,
\begin{equation}
 {J}\left( w \right) = {w^T}\left( {{S_b} - \gamma  {S_w}} \right)w = {w^T}XL{X^T}w.
\end{equation}
Here, $ \gamma $ indicates the trade-off parameter to balance between-class and within-class scatters, and $ L $ refers to the graph Laplacian of MMC \cite{Cheng11DMMC}. Obviously, the solution to such objective can be reached via eign-decomposition of $ J $, such as 
\begin{equation}
 \left( {{S_b} - \gamma {S_w}} \right)w  = XL{X^T} w = \lambda w. 
\end{equation}
Compared with traditional LDA framework, it is able to avoid rank calamity of scatters and exceed the restricted bounding of the category of samples.
Nevertheless, it is hardly to be adopted to high-dimensional data, as eigen analysis of large matrix usually leads to overflow of memory. In light of kernel-view idea, we proposed a direct approach to efficient discriminant analysis \cite{Cheng11DMMC}, and the whole procedure is given in Algorithm 1. 
% ------------------------------------------- 
\begin{algorithm} 
	\caption{The proposed MMC algorithm} 
	\KwIn{Given data points $ X \in {R^{d \times n}} $ in $ c $ classes, the desired reduced  dimensionality $ r $. } 
	\KwOut{ Projective directions, $ {w_i}, i = 1, 2, \cdots ,r $. } 
	1. Calculate between-class scatter $ S_b $ and within-class scatter $ S_w $ with given data;	\\ 
	2. \If{The dimensionality of samples $ < \theta $} 
	{
		2.1 Perform spectral decomposition on $ S_b - \gamma S_w $, and obtain projective directions $ w_i, i = 1, 2, \cdots, r $.	\\
		\Else
		{
			2.2 Construct the sample kernel matrix $ K = {X^T}X $, perform spectral decomposition on $ KLK $, and obtain $ {E^T}KLKE = \Lambda $;	\\ 
			2.3 Calculate the SVD of $ XE{\Lambda ^{ - \frac{1}{2}}} $, and obtain orthogonal matrix $ U $ and singular matrix $ {s_i},i = 1,2, \cdots ,n $;	\\ 
			2.4 Set columns of $ U $ to $ w_i, i = 1, 2, \cdots, r $ in reverse order.	
		} 
	}
\end{algorithm} 
% -------------------------------------------

Though MMC involves the similar discriminant scatters with LDA, the proposed direct approach is quite distinctive compared with traditional ideas. To avoid extra branches of execution, an dimensional threshold is added in MMC. If the sample dimension is less than the given threshold, standard procedure of MMC would be proceeded. On the contrary, an efficient calculational idea would be adopted, and sample kernel $ K=X^T X $ is constructed to reach the kernel scatter. As a consequence, the original MMC problem is transformed to 
\begin{equation}
 J\left( e \right) = {e^T}{X^T}\left( {{S_b} - \gamma {S_w}} \right)Xe = {e^T}KLKe,
\end{equation}
where $ K $ denotes the sample kernel, and $ e $ is the resulting orthogonal directions of MMC. It is noticeable that, the size of decomposed matrix $ \mathds{R}^{d \times d} $ is reduced to $ \mathds{R}^{n \times n} $. Then, the final results can be obtained via calculational tricks of matrix decomposition.

% ------------------------------------------- 
\begin{algorithm} 
	\caption{The RMMC algorithm} 
	\KwIn{Given data points $ X \in {R^{d \times n}} $ in $ c $ classes, the desired reduced dimensionality $ r $, the dimensional threshold $ \theta $ for efficient calculation, and the number of selected samples $ t $. } 
	\KwOut{ Projective directions, $ {w_i}, i = 1, 2, \cdots ,r $. } 
	1. Calculate between-class scatter $ S_b $ and within-class scatter $ S_w $ with given data;	\\ 
	2. \If{The dimensionality of samples $ < \theta $}
	{ 
		2.1 Perform spectral decomposition on $ S_b - \gamma S_w $, and obtain projective directions $ w_i, i = 1, 2, \cdots, r $. 
	
		\Else
		{
			2.2 Randomly select $ t $ samples $ A = \left[ {{x_{a1}}, {x_{a2}}, \cdots ,{x_{at}}} \right] $ from whole data, construct the sample kernel matrix $ M = {A^T}X $, and discriminant scatter $ MLM^T $.	\\ 
			2.3 Follow the similar steps of MMC, and obtain $ w_i, i = 1, 2, \cdots, r $.	\\
%			2.2 Randomly select $ t $ samples $ A = \left[ {{x_{a1,}}{x_{a2}}, \cdots ,{x_{at}}} \right] $ from whole data, and construct the sample kernel matrix $ M = {A^T}X $, perform spectral decomposition on $ MLM^T $, and obtain $ {E^T}MLM^T E $.	\\ 
%			2.3 Calcualte the SVD of $ XE{\Lambda ^{ - \frac{1}{2}}} $, and obtain orthogonal matrix $ U $ and singular values $ {s_i},i = 1,2, \cdots ,t $;	\\ 
%			2.4 Set columns of $ U $ to $ w_i, i = 1, 2, \cdots, r $ in reverse order.	\\
		} 
	} 
\end{algorithm} 
% -------------------------------------------

The most distinguishing points mainly come from step 3 and 4. In step 3, the SVD is proceeded on $ XE{\Lambda ^{ - \frac{1}{2}}} $, namely, 
\begin{equation}
 XE{\Lambda ^{ - \frac{1}{2}}} = U S V^T,
\end{equation}
which is different from tranditional LDA that generally do similar operation on within-class scatter $ S_w $. Furthermore, the obtained orthogonal vectors $ {u_i},i = 1,2, \cdots ,n $ is sorted in descending order corresponding to discriminant power, though they are actually adhere to the largest singular values. Thereafter, the final projective directions $ w_i $ needs to be reverse vectors of $ u_i $ in step 4. In addition, a dimensional threshold is absorbed into original MMC for adaptive dimensionality reduction. That is, the efficient calculation approach would be referred if dimensionality of original data is larger than given threshold.

It is demonstrated that, the proposed MMC method is competent to deal with linear supervised learning in general, while calculational efficiency is preserved. The computational cost mainly depends on $ O \left( {{n^3}} \right) $ for spectral decomposition, compared with $ O\left( {{d^3}} \right) $ of original MMC. For convenience, such approach is called MMC directly in this context. Nevertheless, there also exists some exceptions that $ n $ is still large for direct calculation of big data, and efficiency is unable to be reached. In terms of this limitation, an improved MMC is designed in this work to make supplement of the previous work of ours. The main improvement refers to construction of sample kernel in MMC, and a subset of whole samples are picked up to form the kernel matrix for following step \cite{Cheng15CRH}. Suppose that there are $ t $ samples are selected, the whole procedure is summarized as random MMC (RMMC) in Algorithm 2. 
% -------------------------------------------------------------------------------------- 
Obviously, the computational complexity reduces to $ O\left( {{t^3}} \right) $ in this improved procedure, and the theoretical basis can be derived. 
%As the range subspace of MMC is determined by whole data samples, all discriminative information can be preserved in step 2. But it is noticeable that, the finally desired projective directions are restricted by $ r $, and there are still partial features from full discriminant power are preserved. As a consequence, MMC can be approximated with subspace spanned by a subset of random samples, just like some random methods do, e.g., CRH \cite{Cheng15CRH} or CUR decomposition. Though certain useful information is out of reach in theory, it is surprised to learn the discriminant power has never been much released in performance.

% ------------------------------------------- 
\begin{algorithm} 
	\caption{The single LMMC algorithm} 
	\KwIn{Given data points $ X \in {R^{d \times n}} $ in $ c $ classes, the desired reduced dimensionality $ r $, the dimensional threshold $ \theta $ for efficient calculation, the median dimension $ m $, and the number of selected samples $ t $. } 
	\KwOut{ Projective directions, $ {w_i}, i = 1, 2, \cdots ,r $. } 
	1. Calculate between-class scatter $ S_b $ and within-class scatter $ S_w $ with given data;	\\ 
	2. \If{The dimensionality of samples $ < \theta $}
	{ 
		2.1 Perform spectral decomposition on $ S_b - \gamma S_w $, and obtain projective directions $ w_i, i = 1, 2, \cdots, r $. \\
		\Else
		{
			2.2 Construct the random projection matrix $ P \in \mathds{R}^{d \times g} $, randomly select $ t $ samples $ A = \left[ {{x_{a1}}, {x_{a2}}, \cdots ,{x_{at}}} \right] $ from whole data, and calculate the projection matrix $ B = P P^T $.	\\
			2.3 Construct the sample kernel matrix $ M = {A^T} B X $, and discriminant scatter $ MLM^T $.	\\ 
			2.3 Follow the similar steps of MMC, and obtain $ w_i, i = 1, 2, \cdots, r $.	\\ 
%			2.3 Construct the sample kernel matrix $ M = {A^T} B X $, perform spectral decomposition on $ MLM^T $, and obtain $ {E^T}MLM^T E $.	\\ 
%			2.3 Calcualte the SVD of $ XE{\Lambda ^{ - \frac{1}{2}}} $, and obtain orthogonal matrix $ U $ and singular values $ {s_i},i = 1,2, \cdots ,t $;	\\ 
%			2.4 Set columns of $ U $ to $ w_i, i = 1, 2, \cdots, r $ in reverse order.	\\ 
		} 
	} 
\end{algorithm} 
% -------------------------------------------
\section{Adaptive Learning of MMC}
%In this section, several developments of MMC are given to address diverse categories of handled data in compliated scenarios.
As progress of information technology, there are diversity of handled data categories and application scenarios. The goal of adaptive learning is to exploit the intrinsic patterns of data with different analysis demands in a unified framework as possible.

% ------------------------------------------- 
\begin{algorithm} 
	\caption{The  2D$ ^2 $MMC algorithm} 
	\KwIn{Given each 2D data $ x_i \in \mathds{R}^{\alpha \times \beta}, i = 1, 2, \cdots, n $, in $ c $ classes, the desired reduced dimensionality $ l $ and $ r $, the dimensional threshold $ \theta $ for efficient calculation. } 
	\KwOut{ Bi-directional projective directions, $ {p_i}, {q_j}, i = 1, 2, \cdots, l, j = 1, 2, \cdots, r $. } 
	1. Calculate 2D scatters of column and row (left and right) directions: $ S_{bl} \in \mathds{R}^{d_1 \times d_1} $, $ S_{br} \in \mathds{R}^{d_2 \times d_2} $, $ S_{wl} \in \mathds{R}^{d_1 \times d_1} $, $ S_{wr} \in \mathds{R}^{d_2 \times d_2} $ with given 2D data;	\\
	2. Calculate row projective directions:	\\
		\If{The length of height $ d_1  <  \theta $}
		{
			2.1 Perform spectral decomposition on $ S_{bl} - \gamma S_{wl} $, and obtain projective directions $ p_i, i = 1, 2, \cdots, l $.	\\
			\Else
			{
				2.2 Construct the randomly reduced scatters with lower height as done in MMC.	\\
				2.3 Follow the similar steps of MMC, and obtain $ p_i, i = 1, 2, \cdots, l $.	\\
			}
		}
	3. Calculate column projective directions:		\\
		\If{The length of width $ d_2  <  \theta $}
		{
			3.1 Perform spectral decomposition on $ S_{br} - \gamma S_{wr} $, and obtain projective directions $ q_i, i = 1, 2, \cdots, r $.	\\
			\Else
			{
				3.2 Construct the randomly reduced scatters with shorter width as done in MMC.	\\
				3.3 Follow the similar steps of MMC, and obtain $ q_i, i = 1, 2, \cdots, r $.	\\
			}
		}

\end{algorithm} 
% -------------------------------------------
\subsection{Layered MMC and 2D MMC}
%In the literature, the layered structures have been widely adopted to discriminative learning. 
%With a random projection layer with certain assumptions, e.g., guassian distribution, \cite{Cheng16RM}
With a multi-layer structure, it is believed that the hidden features can be exploited by enlarging original ones from data~\cite{Huang12ELM}. 
%Nevertheless, such consideration takes similar convertion as direct MMC with inner-steps of kernel framework. 
More specifically, a median layer is added to transform each data into a much high-dimensional space, and reduced in following steps with general feature learning. Supposed that there is a given data $ x $, the transformation can be formalized as
\begin{equation}
 x \mapsto h\left( x \right), \mathds{R}^d \mapsto \mathds{R}^g 
\end{equation} 
where $ h\left( \cdot \right) $ denotes the data transformation from original space with dimensionality $ d $ to a much higher dimensionality $ g $ in general. Obviously, such approach is quite identical with kernel learning framework, and can be conduced as a median learning step, e.g.,  
\begin{equation}
 \begin{array}{ll}
{h\left( x \right) = Bx,}&{B = P{P^T}}.
\end{array}
\end{equation}
Here, $ P \in \mathds{R}^{d \times g} $ denotes the linear transformation directions.
For the more specific scenarios, it can be defined as a collaborative learning combined with a nonlinear mapping  $ f\left( x \right) $ and a linear projection and $ P $, which has been employed in extreme learning machine (ELM) \cite{Huang12ELM}. Surprisingly, it is learned that there is a little disparity among different layered approches for MMC. The whole procedure of layered MMC is given as LMMC algorithm.

% --------------------------------------------------------------------------------------
\begin{algorithm} 
	\caption{The single L2D$ ^2 $MMC algorithm} 
	\KwIn{Given each 2D data $ x_i \in \mathds{R}^{\alpha \times \beta}, i = 1, 2, \cdots, n $, in $ c $ classes, the desired reduced dimensionality $ l $ and $ r $, the dimensional threshold $ \theta $ for efficient calculation. } 
	\KwOut{ Bi-directional projective directions, $ {p_i}, {q_j}, i = 1, 2, \cdots, l, j = 1, 2, \cdots, r $. } 
	1. Calculate 2D scatters of column and row (left and right) directions: $ S_{bl} \in \mathds{R}^{d_1 \times d_1} $, $ S_{br} \in \mathds{R}^{d_2 \times d_2} $, $ S_{wl} \in \mathds{R}^{d_1 \times d_1} $, $ S_{wr} \in \mathds{R}^{d_2 \times d_2} $ with given 2D data;	\\
	2. Construct the random matrix $ P \in \mathds{R}^{h1 \times d1} $ and $ Q \in \mathds{R}^{h2 \times d2} $, and transform orignal scatters into high-dimensional spaces.	\\
	3. Calculate row projective directions:	\\
		\If{The length of height $ d_1  <  \theta $}
		{
			3.1 Perform spectral decomposition on $ S_{bl} - \gamma S_{wl} $, and obtain projective directions $ p_i, i = 1, 2, \cdots, l $.	\\
			\Else
			{
				3.2 Follow the similar steps of RMMC, and perform the economical calculations with randomly select sample data. Obtain $ p_i, i = 1, 2, \cdots, l $.	\\

			}
		}
	4. Calculate column projective directions:		\\
		\If{The length of width $ d_2  <  \theta $}
		{
			4.1 Perform spectral decomposition on $ S_{br} - \gamma S_{wr} $, and obtain projective directions $ q_i, i = 1, 2, \cdots, r $.	\\
			\Else
			{
				4.2 Follow the similar steps of RMMC, and perform the economical calculations with randomly select sample data. Obtain  $ q_i, i = 1, 2, \cdots, r $.	\\

			}
		}

\end{algorithm} 
% --------------------------------------------------------------------------------------
On the other hand, 
there are lots of real-world applications intuitively refer to multi-dimensional media information, e.g., images, videos, which rely on surface of 2D-dimensional space. In order to handle those kinds of data directly, 
some 2D based methods are to make learning succinct, e.g., two-dimensional PCA \cite{Yang04TDPCA}\cite{Kong05G2DPCA}, two-dimensional LDA \cite{Li05TDLDA}\cite{Xiong05TDLDA}\cite{Ye04TDLDA}. In general, 2D raw data $ x \in \mathds{R}^{m \times n} $ is involved to find reflecting information between rows of images, e.g., $ y = x V $. 
%And it is believed that, 2D methlods are also competent to reach the sufficient conditions on Bayes optimal compared with 1D ones \cite{Zheng08vs}. 
As the original 2D methods that calculate the single direction for feature extraction \cite{Yang04TDPCA}\cite{Li05TDLDA}, two-directional $ \& $ two-dimensional methods, e.g., 2D$^2$PCA \cite{Kong05G2DPCA} and 2D$^2$LDA \cite{Xiong05TDLDA}\cite{Ye04TDLDA}, are proposed to address the limitation of single-directional learning.

In terms of this consideration, 2D$ ^2 $MMC is devised as a natural extension of original MMC. The main difference between 2D$ ^2 $MMC and original ones is on the fact that 2D data are referred in construction of scatters, while certain steps need to be modified correspondingly.
For a given 2D data $ x \in \mathds{R}^{d_1 \times d_2} $, it aims to find bi-directional projections $ P \in \mathds{R}^{d_1 \times l} $ and $ Q \in \mathds{R}^{d_2 \times r} $, and yields a smaller 2D data $ y \in \mathds{R}^{l \times r} $, e.g.,
\begin{equation}
 y = P^T x Q.
\end{equation}
The calculation of $ P $ and $ Q $ is mainly based on construction of 2D scatters with respect to MMC, e.g.,
\begin{equation}
 \begin{array}{ll}
{S_{bl}} & =  \sum\limits_{i = 1}^c {{n_i}\left( {m_i^2 - {m^2}} \right){{\left( {m_i^2 - {m^2}} \right)}^T}} \\
{S_{wl}} & = \sum\limits_{i = 1}^c {\sum\limits_{j = 1}^{{n_i}} {\left( {x_{ij}^2 - m_i^2} \right){{\left( {x_{ij}^2 - m_i^2} \right)}^T}} } \\
{S_{br}} & = \sum\limits_{i = 1}^c {{n_i}{{\left( {m_i^2 - {m^2}} \right)}^T}\left( {m_i^2 - {m^2}} \right)} \\
{S_{wr}} & = \sum\limits_{i = 1}^c {\sum\limits_{j = 1}^{{n_i}} {{{\left( {x_{ij}^2 - m_i^2} \right)}^T}\left( {x_{ij}^2 - m_i^2} \right)} } 
\end{array}.
\end{equation}
Here, $ S_{bl} $, $ S_{wl} $, $ S_{br} $ and $ S_{wr} $ indicate the 2D scatters with respect to left and right directions, $ m_i^2 $ and $ m^2 $ denote the 2D data of the $ i $-th intra-class mean and the total mean respectively, and $ x_{ij}^2 $ denotes the $ j $-th 2D data belonging to $ i $-th class. Then, the desired $ P $ and $ Q $ can be obtained with standard process of MMC, e.g., 
\begin{equation}
 \begin{array}{l}
J\left( p \right) = {p^T}\left( {S_{bl} - \gamma S_{wl}} \right)p\\
J\left( q \right) = {q^T}\left( {S_{br} - \gamma S_{wr}} \right)q.
\end{array}
\end{equation}
and the whole procedure is summarized as 2D$ ^2 $MMC algorithm.

Similarly, it is also feasible to employ a dimension threshold to ensure the calculational efficiency, especially if large 2D data are referred, e.g., high-resolution images. The related dimension of directional side is reduced with data kernel if the original length ($ \alpha $ or $ \beta $) is larger than threshold $ \theta $. By an example, this can be done with randomly selected rows (or columns) of 2D sample data. 
% --------------------------------------------------------------------------------------
Furthermore, it is straightforward to extend original 2D methods to layered ones. Due to limited space, only the layered 2D$ ^2 $MMC is discussed here. The main branches are quite similar to LMMC algorithm, that is, row and column projections are transformed into high-dimensional space firstly, and followed by MMC approach. The differences come from the handling of high dimensionality of mapped 2D data, which can also be conducted with solution of RMMC similarly. Instead, only one sample data is generally selected to apply economical calculations, and partial columns (or rows) are employed. The single layered-2D$ ^2 $MMC (L2D$ ^2 $MMC) is summarized in Algorithm 5. As a consequence, multi-layered 2D$ ^2 $MMC can be deduced easily, and is able to be proceeded in hierarchical structures of sequential networks.

\subsection{MMC Network}
%Furthermore, there arise much interests in machine learning with artificial neural networks (ANN) structures. 
Inspired by convolution neural network (CNN), PCA Network (PCA-Net) is able to learn classification features of images with a very simple deep learning network \cite{Chan18PCANet}. Nevertheless, the composition of PCA-Net is only the very basic data processing components: cascaded principal component analysis (PCA), binary hashing, and block-wise histograms. In the PCA-Net architecture, PCA is adopted to learn multistage filter banks, followed by simple binary hashing and block histograms for indexing and pooling. With easy and efficient implementation, PCA-Net has been widely employed to learn deep features of objects. Obviously, it is straightforward to extend PCA-Net to MMC, e.g., MMC Network (MMC-Net).
%, and actually, the main differences between them mainly rely on learned cascaded filters are from PCA or MMC.

For each image data $ x_i \in \mathds{R}^{m \times n} $, an image patch is taken around each pixel with size of $ k_1 \times k_2 $ as the manners of local binary patterns (LBP) \cite{Ojala02LBP}. As a consequence, there are $ m \times n $ vectorized patches picked up from $ x_i $, i.e., $ {x_{i,1}},{x_{i,2}}, \cdots ,{x_{i,mn}} \in {R^{{k_1}{k_2}}} $. Assume that the mean-removed patches of each image is indicated by $ \widetilde X $. Then, the class mean and intra-class scatter of $k$-th category with $ n_k $ images can be defined as
\begin{equation}
 \begin{array}{l}
{m_k} = \frac{1}{{{n_k}}}\sum\limits_{i \in {c_k}} {\widetilde {{X_i}}} \\
{S_\phi } = \frac{1}{{{n_k}}}\sum\limits_{i \in {c_k}} {\left( {\widetilde {{X_i}} - {m_k}} \right)} {\left( {\widetilde {{X_i}} - {m_k}} \right)^T}.
\end{array}
\end{equation}
Similarly, the inter-class scatter of image patches belonging to different categories can be defined as
\begin{equation}
 {S_\psi } = \frac{1}{{{n_c}}}\sum\limits_{c = 1}^c {\left( {{m_k} - m} \right){{\left( {{m_k} - m} \right)}^T}}, 
\end{equation}
where $ m $ indicates the mean of class means. With the repeated PCA-Net stage, the output is composed of hashing and histogram of input images.

%In the second stage, it is to almostly repeat the same process as the first stage. And the output stage is composed of hashing and histogram of input images. 

% --------------------------------------------------------------------------------------
\begin{figure*}
\centering
\subfigure[]{
\includegraphics[width=0.54\textwidth]{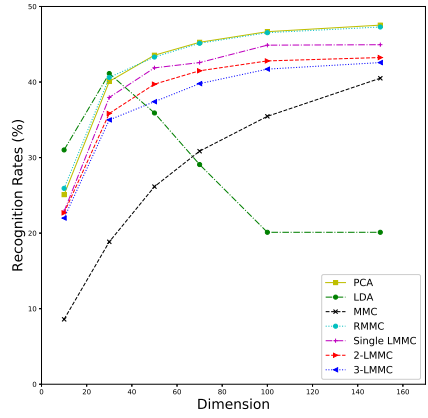}}
\subfigure[]{
\includegraphics[width=0.54\textwidth]{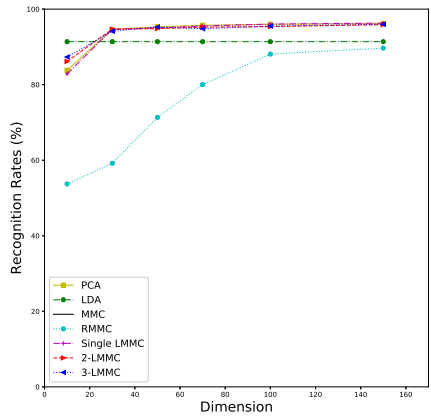}}
\subfigure[]{
\includegraphics[width=0.54\textwidth]{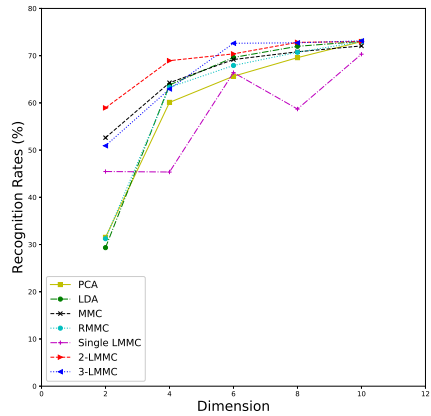}}
\caption{The results of different methods on three data sets. (a) Experimental results on SUN database. (b) Experimental results on MNIST database. (c) Experimental results on STL-10 data set. }
\end{figure*}
% --------------------------------------------------------------------------------------
\section{Experiments}
In this section, several experiments are performed to evaluate the performance of proposed MMC methods\footnote{The implementations are available at: http://mch.one/resources}. First of all, the ability of linear feature extraction is tested, and three data sets, namely SUN scene categorization database\footnote{http://vision.princeton.edu/projects/2010/SUN} \cite{Xiao16SUN}, MNIST digit database\footnote{http://yann.lecun.com/exdb/mnist} \cite{Cun98MNIST}, and STL-10 data set \cite{Coates11STL}, are involved. 
%%For each data in MNIST, 784 dimensional gray features are used to describe its visual handwritten patterns. 
%The SUN scene categorization database contains 108,755 images distributed in 397 well-sampled categories. The number of images varies across categories, but there are at least 100 images per category \cite{Xiao10SUN}\cite{Xiao16SUN}. The MNIST uses 784 dimensional gray features to desceibe its visual handwritten patterns of each data \cite{Cun98MNIST}. The STL-10 dataset uses higher resolution (96x96) images, and 
%%but allows many fewer training examples (100 per class) compared with testing examples, while providing a large unlabeled training set—thus forcing algorithms to rely heavily %on acquired prior knowledge of image statistics \cite{Coates11STL}. In details, the STL-10 dataset 
%contains 10 object classes with 5,000 training and 8,000 test images \cite{Coates11STL}. 
%%There are 10 pre-defined folds of training images, with 500 images in each fold.
%The ALOI database is a color image collection of one-thousand small objects, recorded for scientific purposes. It recorded over a hundred images of each object with varied viewsing angle, illumination angle, and %illumination color for each object, yielding a total of 100,250 images for the collection.

In the SUN database 
%\cite{Xiao10SUN}
\cite{Xiao16SUN}, 
the deep features of each image are extracted by keras toolkit\footnote{https://keras.io} with pre-learned VGG-16 model of imagenet, and a 512 dimensional feature is obtained to describe each image. Among all categories, random 100 classes are selected to be employed in experiments, and random half images of each class are used for training and testing, respectively. Among image data of each digit in the MNIST data set \cite{Cun98MNIST}, 
2,000 images in training data are randomly selected to form training set, while 500 images in testing data are used for testing stage. As a consequence, the total training set are organized by 20,000 images, while testing set contains 5,000 images. Furthermore, the simple sparse coding features of MNIST data are adopted to make an improvement for classification \cite{Labusch08SMHPDR}. For STL-10 data set 
\cite{Coates11STL}, 
the deep representation of each image with target coding is employed in experiments \cite{Yang15DRL}, and a 255 dimensional feature is adopted for each data. Similarly, separate 2,000 and 500 data from each training and testing categories are randomly selected to be training and testing set correspondingly.
% --------------------------------------------------------------------------------------
\begin{figure*}
\centering
\subfigure[]{
\includegraphics[width=0.56\textwidth]{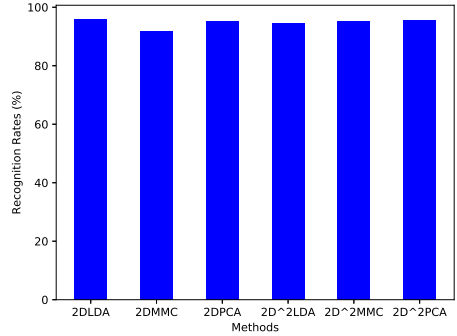}}
\subfigure[]{
\includegraphics[width=0.56\textwidth]{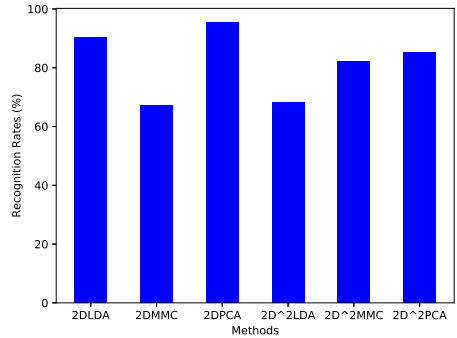}}
\caption{The results of different 2D methods on three data sets. (a) Experimental results on ALOI database. (b) Experimental results on MNIST database. }
\end{figure*}
% --------------------------------------------------------------------------------------

The results on different data sets are shown in Fig. 1. For simplicity, the amount of randomly selected samples are set to be double quantity of reduced dimension $ r $ for RMMC algorithm. 
%Furthermore, different LMMC algorithms with diverse multi-layers are adopted, e.g., single-layer MMC, 2-LMMC, and 3-LMMC, and linear transformation are involved for %simplicity. 
In terms of results, PCA can present stable results for SUN and MNIST data sets, but cannot learn discriminative information with unsupervised features. Similarly, LDA is only up to discriminant analysis for two data sets. The results of RMMC gets close to the best ones, while MMC is incompetent to pattern analysis compared with other linear methods for SUN database. Especially, MMC is hardly to be proceeded for MNIST in our experiments with both high dimensionality and large sample amount, which can be accomplished by RMMC instead. Furthermore, RMMC is able to reach results approximate to MMC in most cases, but much more efficiency can be preserved. For the layered MMC algorhtms, stable performance are still available, and deeper layers lead to better recogniton performance except for SUN data set.

In the second experiment, the discriminant ability of 2D features are evaluated, while the ALOI\footnote{http://aloi.science.uva.nl} and MNIST databases are involved. In the ALOI data set, 
\cite{Geusebroek05ALOI}, 
the whole data are combined while the original order of data is disordered, and then a subset of 50 categories are randomly selected to be involved into experiments. For each category of object, separate 18 and 54 images are randomly picked up to form a small training set compared with testing set of remaining images. For each digit of MNIST, 2,000 and 500 images from training and testing sets are randomly selected to be 2D data, respectively. 
%The experimental results of different 2D methods are shown in Fig. 2.
Since most methods give the close results in different dimensions, the bar chart is adopted to illustrated the results in Fig. 2.

% --------------------------------------------------------------------------------------
\begin{figure*}
\centering
\subfigure[]{
\includegraphics[width=0.56\textwidth]{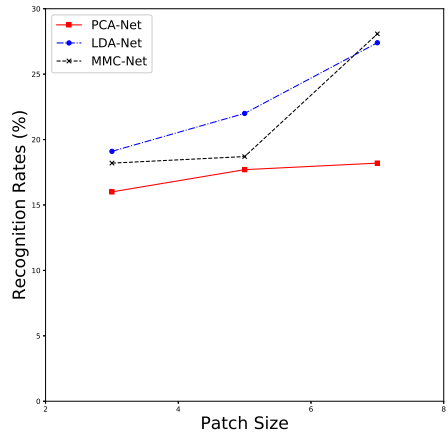}}
\subfigure[]{
\includegraphics[width=0.56\textwidth]{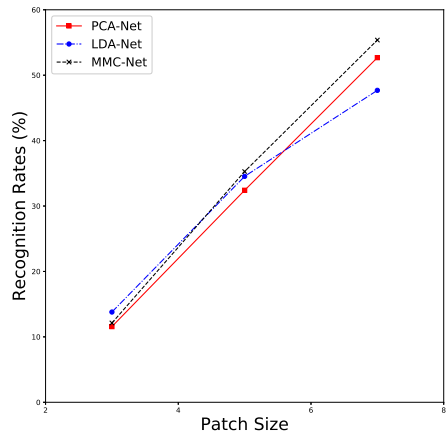}}
\caption{The results of different 2D methods on three data sets. (a) Experimental results on MNIST data set. (b) Experimental results on ALOI data set. }
\end{figure*}
% --------------------------------------------------------------------------------------

As experimental results shown, the discriminant ability of different 2D methods are quite close to each other for ALOI data set. And the best result is contributed by 2DLDA with 95.89\%, followed by 95.63\% of 2D$ ^2 $PCA. Nevertheless, it is noticeable that, 2DMMC can obtain close result 91.7\%, and 2D$ ^2 $MMC is able to reach 95.37\%. In other words, MMC methods can attain similar performance to other methods. For MNIST database, 2DPCA presents the best result of 95.44\%, and other methods are hardly to reach above 90\%. On the contrary, the results from 2D$ ^2 $LDA and 2DMMC are pessimistic among all algorithms, but hopefully, 2D$ ^2 $MMC can still get close recognition results to 2D$ ^2 $PCA method.

In the third experiment, different dimensionality reduction methods are adopted to learn deep neural network structures, i.g., PCA, LDA and MMC Networks, are involved. Two data sets, MNIST and ALOI, are involved in this experiment. For each digit in MNIST, 100 images are randomly selected from training and testing sets, while random 30 categories are selected from ALOI data set with reshaped size of 30$ \times $30. To reduce calculational complexity, three stages only are employed to learn the filter banks, and number of filters are set to be eight. With different size of patch sizes, it is able to disclose the intrinsic affection on subspace neural networks. The experimental results of patch sizes of 3, 4, 5 on two data sets are shown in Fig. 3. In terms of the results, there are few differences among three methods, and both LDA-Net and MMC-Net can reach better results in the stage of dimensionality reduction compared with PCA-Net. Furthermore, it seems that quite similar performance can be obtained with small patch sizes.

\section{Conclusion}
As a classical learning method, MMC is quite popular in various fields of data mining and pattern analysis, as well as its ubiquitous applications in intelligent computing.
In this work, 
%the intrinsic property of maximum margin criterion is discussed, and further developments are devised to meet several scenarios of adaptive learning. 
%In terms of kernel-view process, 
a direct MMC approach is given firstly, and then several variants of MMC are degisned for adaptive learning, i.g., random MMC, layered MMC, 2D$ ^2 $ based MMC. 
%With respect to MMC framework, random MMC is competent to learn discriminative information with subspaces spanned by randomly selected samples. As layered %structures has become familiar with feature extraction applications, a layered MMC is proposed to learn detailed information with multi-layered structure. %Furthermore, a 2D$ ^2 $ based MMC is given to handle the 2D data naturally, and bi-directional diretions of projections are able to learn row and column %shrinkage with discriminant power compared with single-directional 2D methods. Particularly, deep neural networks have been proved to present better %performance for machine learning, and widely adopted to large-scale data mining. 
Inspired by PCA Network, a MMC network method is proposed to make simple deep learning applicable. Experiments on several data sets demonstrate comparable performance of proposed methods for applications of different categories of data types, and it is compenent to learn the associated patterns for adaptive recognition.\\
\\
\textbf{Acknowledgements.}
The authors would like to thank Universit{\"a}t zu L{\"u}beck for sparse coding data set of MNIST, and the Chinese University of Hong Kong for target coding data set of STL-10. The corresponding author of this work is Dr. Miao Cheng.

%\section*{Acknowledgements}
%The authors would like to thank Universit{\"a}t zu L{\"u}beck for sparse coding data set of MNIST, and the Chinese University of Hong Kong for target coding data set of STL-10. The corresponding author of this work is Dr. Miao Cheng.

% ******************************************************************************************************
%\bibliographystyle{IEEEtran}
%\bibliography{mmc}

\end{document}